\documentclass[11pt,a4paper]{article}
\usepackage{times,latexsym}
\usepackage{url}
\usepackage[T1]{fontenc}

\usepackage[acceptedWithA]{tacl2021v1}

\usepackage{xspace,mfirstuc,tabulary}

\newif\iftaclinstructions
\taclinstructionsfalse 
\iftaclinstructions
\renewcommand{\confidential}{}
\renewcommand{\anonsubtext}{(No author info supplied here, for consistency with
TACL-submission anonymization requirements)}
\newcommand{\instr}
\fi

\iftaclpubformat 

\else

\fi

\usepackage{graphicx}
\usepackage{color}
\usepackage{xcolor}
\usepackage{booktabs}
\usepackage{multirow}
\usepackage{hyperref}
\usepackage{caption}
\usepackage{amsmath}
\usepackage{subcaption}
\usepackage{amsfonts, amssymb}
\usepackage{colortbl}
\usepackage{CJKutf8}
\usepackage{tabularx}
\usepackage{enumitem}

\definecolor{purple}{RGB}{128, 0, 128}
\definecolor{LightRed}{rgb}{1,0.92,0.92}
\definecolor{LightOrange}{rgb}{1,0.95,0.88}
\definecolor{LightYellow}{rgb}{1.0,1.0,0.84}
\definecolor{LightGreen}{rgb}{0.9,1.0,0.88}
\definecolor{LightCyan}{rgb}{0.9,1,1}
\definecolor{LightBlue}{rgb}{0.9,0.94,1}
\definecolor{LightIndigo}{rgb}{0.92,0.9,1}
\definecolor{LightMagenta}{rgb}{0.96,0.86,1}
\definecolor{DirtyWhite}{rgb}{0.96,0.96,0.96}

\DeclareSymbolFont{extraup}{U}{zavm}{m}{n}
\DeclareMathSymbol{\varheart}{\mathalpha}{extraup}{86}
\DeclareMathSymbol{\vardiamond}{\mathalpha}{extraup}{87}
\DeclareMathSymbol{\varclubsuit}{\mathalpha}{extraup}{88}

\title{ChiMed 2.0: Advancing Chinese Medical Dataset in Facilitating\\Large Language Modeling}

\author{
    Yuanhe Tian$^{\varheart*}$, \hspace{0.1cm}
    Junjie Liu$^{{\spadesuit*}}$, \hspace{0.1cm}
    Zhizhou Kou$^{{\spadesuit}}$, \hspace{0.1cm}
    Yuxiang Li$^{{\spadesuit}}$, \hspace{0.1cm}
    Yan Song$^{{\spadesuit}}$
    \\
    $^{\varheart}$University of Washington \hspace{0.1cm}
    $^{\spadesuit}$University of Science and Technology of China \\
    $^{\varheart}$\texttt{yhtian@uw.edu} \hspace{0.1cm}
    $^{\spadesuit}$\texttt{ljj19937347730@mail.ustc.edu.cn} \hspace{0.1cm} \\
    $^{\spadesuit}$\texttt{kouzhizhou@mail.ustc.edu.cn} 
    \hspace{0.1cm}
    $^{\spadesuit}$\texttt{liyuxiang2004@mail.ustc.edu.cn} \\
    $^{\spadesuit}$\texttt{clksong@gmail.com}
}

\date{}

\begin{document}
\maketitle

\renewcommand{\thefootnote}{\fnsymbol{footnote}}
\footnotetext[1]{Equal contributions.}
\renewcommand{\thefootnote}{\arabic{footnote}}

\begin{abstract}
\textcolor{black}{
Building high-quality data resources is crucial for advancing artificial intelligence research and applications in specific domains, particularly in the Chinese medical domain.
Existing Chinese medical datasets are limited in size and narrow in domain coverage, falling short of the diverse corpora required for effective pre-training.  
Moreover, most datasets are designed solely for LLM fine-tuning and do not support pre-training and reinforcement learning from human feedback (RLHF).
In this paper, we propose a Chinese medical dataset named ChiMed 2.0, which extends our previous work ChiMed, and covers data collected from Chinese medical online platforms and generated by LLMs.
ChiMed 2.0 contains 204.4M Chinese characters covering both traditional Chinese medicine classics and modern general medical data, where there are 164.8K documents for pre-training, 351.6K question-answering pairs for supervised fine-tuning (SFT), and 41.7K preference data tuples for RLHF.
To validate the effectiveness of our approach for training a Chinese medical LLM, we conduct further pre-training, SFT, and RLHF experiments on representative general domain LLMs and evaluate their performance on medical benchmark datasets.
The results show performance gains across different model scales, validating the dataset’s effectiveness and applicability.\footnote{The information about the data source is available at \url{https://github.com/synlp/ChiMed-2.0}.}
}
\end{abstract}


\section{Introduction}

\textcolor{black}{
Recent years have witnessed remarkable progress in large language models (LLMs) \cite{ouyang2022training,taori2023alpaca,touvron2023llama,yang2025qwen3}, which achieve outstanding performance in many downstream tasks \cite{park2023generative,tian2024dialogue,abbasiantaeb2024let}.
These models are increasingly adopted across professional domains such as legal analysis, financial forecasting, and scientific literature review \cite{liu2023fingpt,zhou2024lawgpt,tian-etal-2024-chimed}.
In particular, LLMs are applied to medical domain tasks such as medical report generation, clinical decision support, and patient triage \cite{yuan2024continued,monajatipoor2024llms,tian2024diffusion,tian2025computed,dai2025zeus}.
Within this area, Chinese-language applications are critical, given the high demand for online medical consultation services in China.
Training such domain-specific LLMs requires large-scale, high-quality datasets that capture medical terminology, patient narratives, and clinical guidelines.
However, existing Chinese medical corpora are limited in size and scope, often focusing on general health topics without covering specialized clinical scenarios.
Thus, building a comprehensive Chinese medical dataset that integrates broad knowledge sources and detailed clinical content is essential.}

\begin{table*}[t]
\centering
\setlength{\tabcolsep}{4pt}
\begin{tabular}{lcccr}
\toprule
\textbf{Dataset} & \textbf{Language} & \textbf{Source} & \textbf{Type} & \multicolumn{1}{c}{\textbf{Size}} \\
\midrule
ChiMed \cite{tian2019chimed} 
& CH & Online Platform  & QA & 52.21M  \\
PubMedQA \cite{jin2019pubmedqa} & EN & PubMed Summary & QA & 14.96M \\
MedicationQA \cite{abacha2019bridging} & EN & Online Platform & QA & 0.06M \\
MedQA \cite{jin2021disease} & CH/EN & Medical Exam & QA & 6.76M  \\
MedMCQA \cite{pal2022medmcqa} & EN & Medical Exam & QA & 17.30M \\ 
CMB \cite{wang2023cmb} & CH & Question Database & QA & 12.03M \\
CMExam \cite{liu2023benchmarking} & CH & Medical Exam & QA & 12.72M \\
PMC-Patients\cite{zhao2023large} & EN & Clinical Report & Patients Summary & 109.18M \\
cMtMedQA \cite{yang2024zhongjing} & CH & Online Platform & Dialogue &  6.57M \\
MedBullets \cite{chen2025benchmarking} & EN & Open Datasets & QA & 0.18M \\
\midrule
ChiMed 2.0 (this work) & CH & Online Platform & Doc, QA, PD & 204.40M
\\
\bottomrule
\end{tabular}
\caption{
\textcolor{black}{
Summarization of existing representative datasets for Chinese (CH) and English (EN) in the medical domain. The size is illustrated in terms of the number of tokens (Character tokens for Chinese and word tokens for English).
The ``Doc'', ``QA'', and ``PD'' are the abbreviations for ``documents'', ``question answering'', and ``preference data'', respectively.
}}
\label{tab:chinese_med_datasets}
\end{table*}

\textcolor{black}{
Existing medical datasets are predominantly English, where the Chinese resources are scarce.
Table \ref{tab:chinese_med_datasets} summarizes representative Chinese medical datasets, their sources, and sizes \cite{tian2019chimed,jin2019pubmedqa,abacha2019bridging,song-etal-2020-summarizing,jin2021disease,pal2022medmcqa,liu2023benchmarking,wang2023cmb,wang2024assessing,li2024mediq,chen2025benchmarking}.
These datasets originate from diverse sources, including medical licensing exam question banks, online consultation platforms, forum-based QA, and structured knowledge bases.
However, their domain coverage remains confined to QA pairs and encyclopedic snippets, lacking diverse content such as clinical pathways, electronic health records, research literature, and Chinese medicine classics.
Moreover, despite variations in size, the overall scale still falls short of the massive and heterogeneous corpora required for pretraining.
Crucially, existing resources focus on fine-tuning rather than general pretraining and do not adequately address ambiguities in Chinese medical terminology or the specialized knowledge of Chinese medicine.
Thus, there is an urgent need to build a comprehensive Chinese medical dataset that covers a wide range of content and supports pretraining, fine-tuning, and RLHF.
}

\textcolor{black}{
In this paper, we construct a large-scale Chinese medical dataset, named ChiMed 2.0, which extends ChiMed \cite{tian2019chimed} to cover pre-training, fine-tuning, and reinforcement learning from human feedback (RLHF).
The dataset consists of two major components: traditional Chinese medicine classics and modern general medical documents.
The general medical documents include records from online medical sources, forum QA, and patient–doctor dialogues.
Data sources comprise various public medical websites, online consultation platforms, and FAQ databases.
We perform basic preprocessing on the raw data, including noise removal, HTML and URL stripping, and quality filtering based on language model perplexity.
Due to missing information in the raw data, we conduct deep processing with LLMs, including department classification, privacy anonymization, and automated translation of ancient Chinese to modern Chinese in the Chinese medicine classics.
In addition, we also generate question–answer pairs from documents and create preference data to support the SFT and RLHF stages of LLMs
The final dataset contains 204.4M characters, covering diverse medical scenarios and specialized terminology.
This dataset supports LLM pre-training, fine-tuning, and RLHF, so as to further help downstream medical NLP tasks such as diagnostic assistance, clinical decision support, and question answering systems.
We conduct experiments on models of various scales and demonstrate that the dataset improves model accuracy and professionalism scores.
}

Our contributions are summarized as follows:
\begin{itemize}[leftmargin=1em]
  \item We propose ChiMed-2.0, a large‑scale Chinese medical dataset covering pre‑training, supervised fine‑tuning, and reinforcement learning from human feedback, integrating both traditional Chinese medicine classics and modern clinical content.
  \item We design a multi‑stage processing pipeline that consists of deduplication, noise filtering, sensitive content screening, automated ancient‑to‑modern translation, QA pair generation, and preference data construction to ensure high corpus quality and diversity.
  \item We provide detailed dataset statistics and demonstrate, through experiments on CMMLU and CEval benchmarks, that models trained on ChiMed-2.0 achieve consistent performance gains across different scales.
  \item We discuss potential broader applications of ChiMed-2.0 beyond model training, including knowledge graph construction, information retrieval benchmarking, domain‑specific NLP research, and epidemiological analysis.
\end{itemize}

\section{Related Work}

\textcolor{black}{
Medical domain datasets play a key role in training and evaluating language models.
In the English medical domain, researchers have released several prominent datasets for various tasks.
For example, MIMIC-III \cite{johnson2016mimic} originates from the ICU electronic health records at Beth Israel Deaconess Medical Center, including over 2 million clinical notes and discharge summaries for more than 40,000 patients between 2001 and 2012.
Additionally, MedDialog \cite{zeng-etal-2020-meddialog} gathers 260,000 English patient–doctor multi-turn dialogues, spanning 96 specialties and supporting conversational AI research.
Furthermore, PubMedQA \cite{jin2019pubmedqa} comprises 1,000 expert-annotated yes/no/maybe question–answer pairs drawn from PubMed abstracts, supplemented by over 60,000 unlabeled and 200,000 synthetic examples for reasoning over research findings.
For Chinese medical datasets, availability is growing, yet variations in scale and coverage remain.
For example, CMeIE \cite{guan2020cmeie} offers entity and relation annotations over thousands of clinical case reports across cardiovascular, respiratory, and neurological domains.
CMedQA \cite{cui2020chinese} integrates 120,000 consumer health questions with 226,000 expert answers, aimed at training and benchmarking QA models.
Additionally, ChiMed \cite{tian2019chimed} is a large Chinese QA dataset obtained from online healthcare platforms, where the questions come from patients and answers come from registered doctors.
While these datasets foster domain-specific research, they still face limitations in cross-task generalization, uniform data quality, and ongoing updates.
}

\section{The ChiMed 2.0 Dataset}

\textcolor{black}{
Our dataset contains online medical documents, patient–doctor QA platforms, medical encyclopedias, and traditional Chinese medicine classics, with a framework covering pre-training, fine-tuning, and RLHF stages, integrating both traditional Chinese medicine and general medical texts.  
In the following subsections, we provide a detailed overview of the dataset construction and composition from the perspectives of data collection, data processing, and dataset statistics.  
}

\begin{figure}
    \centering
    \includegraphics[width=1.0\linewidth]{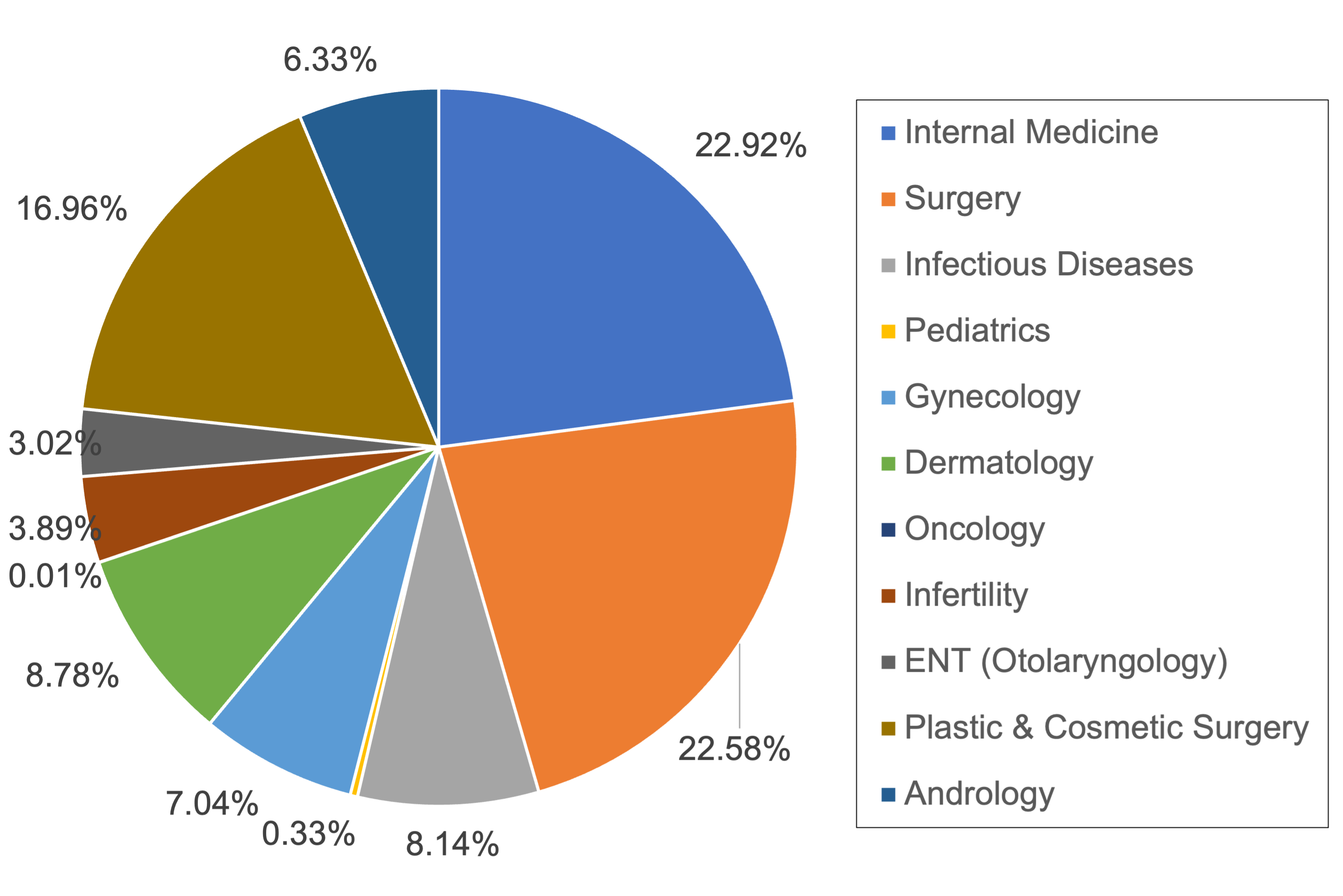}
    \caption{The distribution of the department of the QA pairs from the 39 Ask Doctors.}
    \label{fig:data-distribution}
\end{figure}

\begin{figure*}
    \centering
    \includegraphics[width=0.98\linewidth]{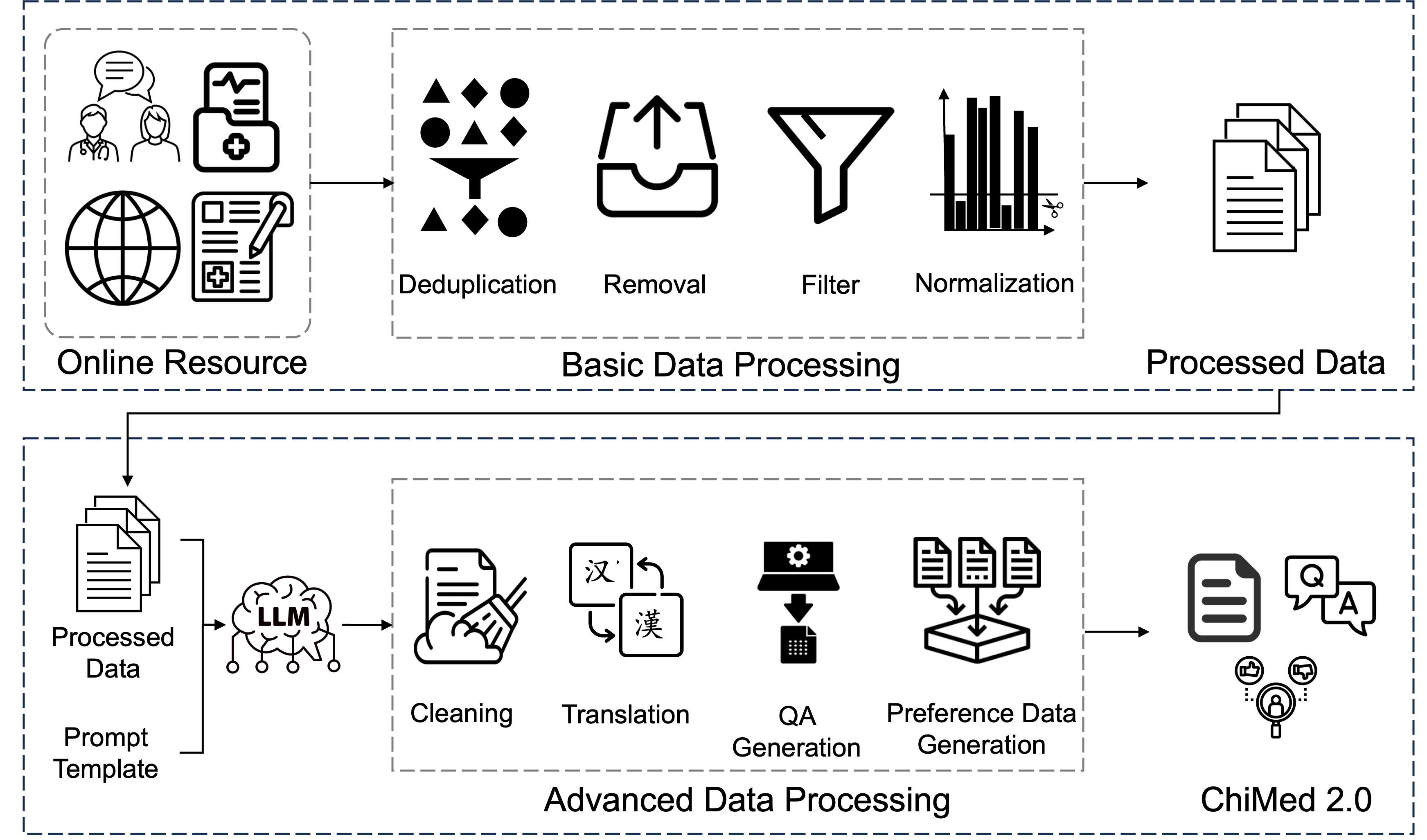}
    \caption{The overall data processing process.}
    \label{fig:processing}
\end{figure*}

\subsection{Data Collection}

\textcolor{black}{
We select multiple authoritative and high-volume websites as data sources, covering traditional Chinese medicine (TCM) and general medical data.
The sites cover clinical guidelines, research reviews, TCM theory, case discussions, and patient consultations.  
Site selection is based on authority and data volume to ensure content quality and diversity.
Specifically, we collect data from the following representative sources:
\begin{itemize}[leftmargin=1em]
     \item Chinese Medicine Dictionary\footnote{\url{www.zydcd.com}}: This nationally authoritative digital platform for TCM knowledge is dedicated to systematically organizing and disseminating the essence of classical Chinese medical culture. Its data is mainly scraped for traditional Chinese medicine, encompassing classical medical texts, clinical case records, and research articles that facilitate classical prescription analysis, diagnostic methodologies, and historical prescription studies.
     \item Traditional Chinese Medicine Database\footnote{\url{www.zhongyoo.com}}: This specialized knowledge service platform for TCM in China is dedicated to delivering precise and comprehensive TCM information retrieval services. Its website content is principally scraped for comprehensive TCM resources, containing medicinal herbs, processing methods, clinical cases, health preservation techniques, patent medicines, acupoints, folk remedies, and classical formulas.  The platform provides advanced analytical capabilities including herb-formula-disease relationship analysis, visual processing techniques, and AI-driven compatibility alerts.
     \item China Health Network\footnote{\url{www.cnkang.com}}: As a nationally renowned comprehensive health portal in China, this platform offers practical services including online consultations with top-tier hospital physicians, health self-assessment tools, and medical care navigation guidance. Its website content is principally scraped for clinical Chinese medicine resources, encompassing classical formula collections and herbal medicine knowledge, while delivering specialized formula applications, herb usage guidelines, and evidence-based therapeutic recommendations.
     \item 39 Ask Doctors\footnote{\url{ask.39.net}}: As a specialized online medical consultation platform, this service hosts thousands of licensed physicians from national public hospitals, delivering round-the-clock virtual consultation services. It contains physician-provided medical consultations and evidence-based treatment suggestions from different departments (the department distribution is presented in Figure \ref{fig:data-distribution}).
     \item 120 Ask Health QA\footnote{\url{www.120ask.com}}: As China's premier health and medical consultation platform, it specializes in delivering professional and convenient online health advisory services to users. 
     The platform aggregates physicians and medical experts from hospitals across all levels nationwide, covering comprehensive clinical departments including internal medicine, surgery, obstetrics/gynecology, pediatrics, and traditional Chinese medicine. 
     Its core content is principally scraped for medical consultation QA resources, capturing real-time doctor-patient interactions and personalized health advice.
\end{itemize}
The raw data contains roughly 221M characters.
}

\begin{table*}[tp]
    \centering
    \scriptsize
    \begin{tabular}{p{15.5cm}}
    \toprule
    \textbf{Cleaning}: \\
    \begin{CJK}{UTF8}{gkai}你是一个医疗内容安全审核助手，专门审核医疗健康相关内容。请仔细分析用户提供的医疗文本，并回答以下问题：\end{CJK} \\
    \begin{CJK}{UTF8}{gkai}
        1. 文本是否包含明显的侮辱、歧视、仇恨言论等有毒内容？\end{CJK} \\
    \begin{CJK}{UTF8}{gkai}
        2. 文本是否包含具体的个人身份信息（如全名、详细地址、电话号码、身份证号、电子邮箱等）？\end{CJK} \\
    \begin{CJK}{UTF8}{gkai}
        注意：医疗术语、疾病描述、治疗建议等内容不被视为有毒内容。\end{CJK} \\
    \begin{CJK}{UTF8}{gkai}
        如果以上任一问题的答案是肯定的，请回答'是'，否则回答'否'，不要生成其他内容。\end{CJK} \\
    \textit{You are a medical content security review assistant, specializing in reviewing medical and health-related content.
    Please carefully analyze the medical text provided by the user and answer the following questions:}\\
    \textit{Does the text contain any obviously toxic content such as insults, discrimination, hate speech, etc.?} \\
    \textit{2. Does the text contain specific personal identity information (such as full name, detailed address, phone number, ID number, email address, etc.)?} \\
    \textit{Note: Medical terms, disease descriptions, treatment suggestions and other such content are not regarded as toxic.} \\
    \textit{If the answer to any of the above questions is affirmative, please answer 'Yes'; otherwise, answer' No '. Do not generate any other content.} \\
    \midrule
    \textbf{Translation}: \\
    \begin{CJK}{UTF8}{gkai}请将以下古文翻译成通俗易懂的现代汉语，保留原文的专业术语和医学概念：<输入文本>\end{CJK} \\
    \textit{Please translate the following ancient text into modern Chinese that is easy to understand, retaining the professional terms and medical concepts of the original text: < Input Text >} \\
    \midrule
    \textbf{QA Generation}: \\
    \begin{CJK}{UTF8}{gkai}你是一个专业的医疗问答生成助手，擅长从医疗文档中提取核心信息并生成相关问题及答案。请根据用户提供的医疗文档，生成一个相关的问题和对应的答案。要求：\end{CJK} \\
    \begin{CJK}{UTF8}{gkai}
1. 问题必须基于文档内容，不能脱离文档 \end{CJK} \\
    \begin{CJK}{UTF8}{gkai}
2. 答案必须直接从文档中得出，不能添加额外信息 \end{CJK} \\
    \begin{CJK}{UTF8}{gkai}
3. 问题和答案都使用中文 \end{CJK} \\
    \begin{CJK}{UTF8}{gkai}
4. 使用以下格式输出：问题：[生成的问题]答案：[从文档中提取的答案]\end{CJK} \\
    \textit{You are a professional medical QA generation assistant, skilled at extracting core information from medical documents and generating relevant questions and answers.
Please generate a relevant question and the corresponding answer based on the medical documents provided by the user.} \\
\textit{Requirements:} \\
\textit{1. The questions must be based on the content of the document and cannot be divorced from it.} \\
\textit{2. The answer must be directly derived from the document and no additional information can be added.} \\
\textit{3. Both the questions and the answers are in Chinese.} \\
\textit{4. Output in the following format:
Question: [Generated question]
Answer: [The answer extracted from the document]} \\
    \midrule
    \textbf{Preference Data Generation}: \\
    (a) Answer Generating \\
    \begin{CJK}{UTF8}{gkai}你是一个医疗问答助手，根据用户的问题提供准确、专业的医疗建议。请确保回答基于医学事实，语言简洁明了，限制在100字以内。
    \end{CJK} \\
    \textit{You are a medical QA assistant, providing accurate and professional medical advice based on users' questions.}\\
    (b) Answer Ranking \\
    \begin{CJK}{UTF8}{gkai}
    你是一个医疗问答质量评估专家。请根据以下标准评估两个回答的质量：\end{CJK} \\
    \begin{CJK}{UTF8}{gkai}
        1. 准确性：回答是否准确，是否符合医学常识\end{CJK} \\
    \begin{CJK}{UTF8}{gkai}
        2. 有用性：回答是否对提问者有帮助\end{CJK} \\
    \begin{CJK}{UTF8}{gkai}
        3. 完整性：回答是否完整覆盖了问题\end{CJK} \\
    \begin{CJK}{UTF8}{gkai}
        4. 简洁性：回答是否简洁明了\end{CJK} \\
    \begin{CJK}{UTF8}{gkai}
        问题：<question>\end{CJK} \\
    \begin{CJK}{UTF8}{gkai}
        回答A: <answer1>\end{CJK} \\
    \begin{CJK}{UTF8}{gkai}
        回答B: <answer2>\end{CJK} \\
    \begin{CJK}{UTF8}{gkai}
        请比较这两个回答，选择质量更高的一个（输出A或B）。
    \end{CJK} \\
    \textit{Please ensure that your answer is based on medical facts, with concise and clear language, and is limited to within 100 words.
    You are an expert in evaluating the quality of medical QA. Please evaluate the quality of the two responses according to the following criteria:} \\
\textit{1. Accuracy: Whether the answer is accurate and in line with medical common sense.} \\
\textit{2. Usefulness: Whether the answer is helpful to the questioner.} \\
\textit{3. Completeness: Whether the answer completely covers the question} \\
\textit{4. Conciseness: Whether the answer is concise and clear} \\
\textit{Question: <question>} \\
\textit{Answer A: <answer1>} \\
\textit{Answer B: <answer2>} \\
\textit{Please compare these two answers and choose the one with higher quality (output A or B).} \\
    \bottomrule
    \end{tabular}
    \caption{Example prompts used to instruct an LLM to perform advanced data processing. All prompts are in Chinese, and the English translation is provided only for reference.
    }
    \label{tab:prompt-example}
\end{table*}

\subsection{Data Processing}

\textcolor{black}{
Our raw data undergoes multi-stage processing, divided into basic and advanced phases, to ensure corpus quality and enrich sample diversity.
Figure \ref{fig:processing} illustrates the overall workflow of the data processing, which is illustrated as follows.
}

\paragraph{Basic Data Processing}

\textcolor{black}{
In the basic processing phase, we firstly apply deduplication according to the URL of items to duplicate items. Next, we use regular expressions to remove HTML tags, CSS styles, URLs, non-text characters, and irrelevant information such as greetings. Then, we concatenate the question and answer of each item, and compute the perplexity (PPL) of each text using Chinese GPT-2 \cite{radford2019language}\footnote{We get the model from \url{https://huggingface.co/uer/gpt2-chinese-cluecorpussmall}.} and remove samples whose PPL values are in the top 1\%. Finally, we normalize text length by discarding samples shorter than 30 tokens.
}

\begin{table*}[tp]
    \centering
    \scriptsize
    \begin{tabular}{p{15.5cm}}
    \toprule
    Trad \\
    \begin{CJK}{UTF8}{gkai}茯苓在加工时将菌核内部的白色切成薄片或是小块，为白。味甘淡，性平，归心、脾、肺、肾经。功能健脾安神、利水渗湿。临床上常用于小便不利、水肿及痰饮等水湿证、脾虚证等。\end{CJK} \\
    \textit{The medicinal herb Poria (Fuling) is processed by slicing or cutting the white interior portion of the fungal sclerotium into thin slices or small pieces, known as Bai Fu Ling. It has a sweet and bland taste, with a neutral property, and it acts on the Heart, Spleen, Lung, and Kidney meridians. Its main functions include strengthening the spleen, calming the mind, promoting diuresis, and resolving dampness. Clinically, it is commonly used to treat syndromes of water retention and dampness, such as difficult urination, edema, phlegm-fluid retention (tan yin), as well as spleen deficiency syndromes.} \\
    \midrule
    Gen \\
    \begin{CJK}{UTF8}{gkai}治疗鼻中隔偏曲的方法有手术和非手术治疗两种，手术治疗是通过手术矫正鼻中隔偏曲，非手术治疗则是通过药物治疗和物理治疗来缓解症状。\end{CJK} \\
    \textit{There are two treatment options for nasal septal deviation: surgical and non-surgical. Surgical treatment involves correcting the deviation through surgery, while non-surgical treatment aims to relieve symptoms through medication and physical therapies.} \\
    \bottomrule
    \multicolumn{1}{c}{(a) Pre-training}\\
    \toprule
    Trad \\
    \begin{CJK}{UTF8}{gkai}Q: 一年四季脚都是凉的，除了运动状态。\end{CJK}\\
    \begin{CJK}{UTF8}{gkai}A: 手脚冰凉，畏寒肢冷，应该同时还伴有腹胀，经常大便偏稀等症状，这个是脾阳虚的表现，脾虚会造成气血不足，也就是西医的血液循环不好，泡点枸杞子也可以，应该有好处，最好是吃中药调理调理。\end{CJK} \\
    Q: \textit{My feet are always cold throughout the four seasons, except during exercise.} \\
    A: \textit{Cold hands and feet, along with sensitivity to cold, are often accompanied by symptoms such as abdominal bloating, loose stools, and frequent diarrhea. These are manifestations of spleen-yang deficiency in traditional Chinese medicine (TCM). Spleen deficiency can lead to insufficient qi and blood, which is similar to poor circulation in Western medicine. Soaking some goji berries (wolfberries) in water may help, and it should be beneficial. However, the best approach is to take Chinese herbal medicine for proper regulation and body conditioning.} \\
    \midrule
    Gen \\
    \begin{CJK}{UTF8}{gkai}Q: 口干的治疗方案是什么?请描述口干的治疗方案。\end{CJK}
    \\
    \begin{CJK}{UTF8}{gkai}A: 口干症的治疗包括病因治疗和对症治疗。对因治疗在明确病因的情况下是最有效的，如药物性口干，通过调整药物及其剂量，可缓解口干。对唾液消耗增加而产生的口干，可通过消除张口呼吸等原因来解决。如果是由于唾液腺实质破坏所引起的口感，如头颈部恶性肿瘤放疗后、舍格伦综合征，目前主要通过对症治疗来缓解口干，减少并发症。\end{CJK} \\
    Q: \textit{What is the treatment plan for dry mouth? Please describe the treatment plan for dry mouth.} \\
    A: \textit{The treatment of xerostomia (dry mouth) includes both etiological treatment and symptomatic treatment. Etiological treatment is most effective when the underlying cause is clear. For example, in cases of drug-induced xerostomia, adjusting the medication or its dosage can help relieve dry mouth. For dry mouth caused by increased saliva consumption, addressing the cause — such as open-mouth breathing — can resolve the issue. If dry mouth results from damage to the salivary gland tissue, such as after radiotherapy for head and neck cancers or in Sjögren's syndrome, symptomatic treatment is currently the main approach. This focuses on relieving symptoms, improving oral hydration, and reducing complications.} \\
    \bottomrule
    \multicolumn{1}{c}{(b) SFT} \\
    \toprule
    Trad \\ 
    \begin{CJK}{UTF8}{gkai}患者：带状疱疹7天左右，想咨询有否中药外敷用药？\end{CJK} \\
    \begin{CJK}{UTF8}{gkai}接受示例：带状疱疹可外用中药如龙胆泻肝汤或紫草油，具有清热解毒、消肿止痛作用。建议配合口服抗病毒药物，并保持患处清洁干燥。若症状加重或出现神经痛，应及时就医；\end{CJK}\\
    \begin{CJK}{UTF8}{gkai}拒绝示例：可外敷炉甘石洗剂或金银花水，缓解瘙痒。若病情加重或持续不退，建议及时就医。\end{CJK} \\
    Patient: \textit{I have been suffering from shingles for about 7 days. I would like to ask if there are any external traditional Chinese medicine applications.} \\
    Accept example: \textit{Herpes zoster can be treated with traditional Chinese medicine externally, such as Longdan Xiegan Decoction or Zicao Oil, which have the effects of clearing heat, detoxifying, reducing swelling and relieving pain. It is recommended to take oral antiviral drugs in combination and keep the affected area clean and dry. If symptoms worsen or neuralgia occurs, medical attention should be sought promptly.} \\ 
    Reject example: \textit{Calamine lotion or honeysuckle water can be applied externally to relieve itching. If the condition worsens or persists without improvement, it is recommended to seek medical attention promptly.} \\
    \midrule
    Gen \\
    \begin{CJK}{UTF8}{gkai}患者：10个月的宝宝不小心吃了一点爽肤露怎么办？\end{CJK}\\
    \begin{CJK}{UTF8}{gkai}\textbf{接受示例}：立即催吐并就医，避免误吸。爽肤露含酒精，可能引起中毒。密切观察宝宝呼吸、意识，及时送医处理；\end{CJK} \\
    \begin{CJK}{UTF8}{gkai}\textbf{拒绝示例}：立即清洗口腔并观察是否有呕吐、嗜睡等异常。若出现不适，及时就医。避免再次接触，保持通风。\end{CJK} \\
    Patient: \textit{What should I do if my 10-month-old baby accidentally eats a little toner?} \\
    Accept example: \textit{Induce vomiting immediately and seek medical attention to prevent aspiration. Toner contains alcohol and may cause poisoning. Closely observe the baby's breathing and consciousness, and send them to the hospital for treatment in a timely manner.} \\
    Reject example: \textit{Immediately clean your mouth and observe if there are any abnormalities such as vomiting or drowsiness. If you feel unwell, seek medical attention promptly. Avoid further contact and keep the area well-ventilated.} \\
    \bottomrule
    \multicolumn{1}{c}{(c) RLHF} \\
    \end{tabular}
    \vspace{-0.3cm}
    \caption{Data examples in ChiMed 2.0, where the examples from the pre-training, SFT, and RLHF for the traditional Chinese medicine (Trad) and general Chinese medicine (Gen) are presented. English translation is provided for readability; it is not a part of the data.}
    \label{tab:data example}
\end{table*}

\begin{table}[t]
    \centering
    \small
    \begin{tabular}{l|l|c|c}
    \toprule
    & & Trad & Gen \\
    \midrule
    \multirow{2}{*}{Pre-train}
    & \# of chars & 55.7M & 2.1M \\
    & \# of documents & 160.8K & 4.0K \\
    \midrule
    \multirow{2}{*}{SFT}
    & \# of chars & 11.2M & 127.1M \\
    & \# of QA pairs & 9.8K & 341.8K \\
    \midrule
    \multirow{2}{*}{RLHF}
    & \# of chars & 2.3M & 6.0M \\
    & \# of data tuples & 11.7K & 30.0K \\
    \midrule
    \midrule
    \multirow{2}{*}{Total} 
    & \# of chars & 69.2M & 135.2M \\
    & \# of instances & 182.3K & 375.8K \\
    \bottomrule
    \end{tabular}
    \caption{The statistics of ChiMed 2.0, where the number of characters, documents, QA pairs, preference data tuples are reported.}
    \label{tab:stat}
\end{table}

\paragraph{Advanced Data Processing}

\textcolor{black}{
The advanced data processing stage consists of sensitive content cleaning, ancient text translation, QA pair generation, and preference data generation, where LLMs are used to perform these processing steps with the prompts illustrated in Table \ref{tab:prompt-example}.
The first stage performs sensitive content cleaning to ensure compliance and protect user privacy, forming the foundation of data security.
We use predefined prompts to instruct an LLM to detect personal identifiers (e.g., names, ID numbers, phone numbers) and sensitive attributes (e.g., race, religion, sexual orientation).
Simultaneously, the prompt directs the model to identify insulting, hateful, or other inappropriate content (e.g., defamation, violent descriptions, and misleading medical claims).
We directly remove any samples flagged by the model as containing private or inappropriate content, ensuring the dataset retains only compliant and safe text.
To control false positive rates and validate screening effectiveness, we perform periodic sampling and manual reviews, further refining our screening strategy.
}

\textcolor{black}{
The second stage addresses TCM document scarcity and the importance of prompt-based training by designing a QA pair generation module to enhance the model’s understanding of document content.  
We craft a specific prompt to drive an LLM to extract key information from chapters and paragraphs and generate QA pairs.  
In total, we generate approximately 220K high-quality QA pairs to serve as rich training samples for subsequent SFT and RLHF stages.  
}

\textcolor{black}{
The third stage addresses the high complexity of classical Chinese in TCM classics, recognizing them as a vital part of the TCM knowledge system but noting that their archaic style hinders model learning.  
We craft a specialized prompt to instruct an LLM in automated translation.  
Through this translation module, we convert numerous classical passages into more readable modern Chinese text.  
The translation results significantly improve corpus readability and enhance the model’s comprehension of ancient TCM knowledge.  
}

\textcolor{black}{
The fourth stage constructs the preference dataset for RLHF, for which a preference data generation module is set up.
We divide this module into two sections: the answer generation section and the quality assessment section.
In the answer generation section, we design a prompt: ``\textit{Provide accurate and professional medical advice based on the following questions}'', to instruct the LLM in generating two different sets of answers to the same question input.
Then, in the quality assessment section, we craft another prompt: ``\textit{Please compare the following two responses and choose the one with higher quality}'', to ask another LLM to evaluate the quality of the two generated answers.
The better answer is regarded as the accepted output and the other one is used as the rejected output, which are used to construct a preference dataset.
The generated preference dataset is able to be used for RLHF training of the LLMs.
}

\textcolor{black}{  
After these advanced processing steps, we obtain high-quality records for subsequent fine-tuning, RLHF training, and evaluation.
}

\subsection{Dataset Properties}

\textcolor{black}{
The dataset is organized into three main parts: pre-training, supervised fine-tuning (SFT), and preference data, each encompassing both TCM and general medical texts.  
In the pre-training part, the data consists primarily of raw medical documents, including clinical guidelines, research reviews, and case reports.  
In the SFT part, the data is composed mainly of generated QA pairs designed to guide the model’s supervised fine-tuning.  
In the preference data part, we include model-generated accept and reject examples for the reinforcement learning from human feedback (RLHF) stage.  
Example data samples are presented in Table \ref{tab:data example}, and overall statistics are reported in Table \ref{tab:stat}.  
This dataset supports LLM pre-training, supervised fine-tuning, RLHF training, and various medical NLP downstream tasks such as diagnostic assistance and clinical decision support.  
}

Beyond its role in LLM pre‑training, fine‑tuning, and RLHF,
ChiMed 2.0 has the potential to support clinical knowledge extraction and ontology development, enabling the creation of detailed Chinese medical knowledge graphs.
It also serves as a satisfactory source to aid in benchmarking and enhancing medical information retrieval and question‑answering systems in real‑world healthcare scenarios.
It has the potential to underpin research in domain‑specific NLP tasks such as named entity recognition, relation extraction, and terminology standardization.
It is possible to further utilize ChiMed 2.0 for epidemiological analysis and public health monitoring, offering insights into symptom distributions, regional variations, and emerging treatment trends.

\begin{table}[t]
\centering
\small
\scalebox{0.95}{
\begin{tabularx}{0.5\textwidth}{l|ccc}
\toprule
 & Pre‑training & Fine‑tuning & RLHF \\
\midrule
Epochs         & 2   & 2   & 1    \\
Batch size     & 64 & 32  & 16   \\
Learning rate  & 1e-5 & 5e-5 & 1e-5 \\
Warmup ratio   & 0.1 & 0.1 & 0.1 \\
\bottomrule
\end{tabularx}
}
\vspace{-0.2cm}
\caption{
\textcolor{black}{
The hyperparameter settings for training our model on ChiMed 2.0.
}
}
\label{tab:hyperparameters}
\end{table}

\section{Experiments}

\subsection{Settings}

In our experiments, we first pre-train models on the constructed dataset, then apply supervised fine-tuning (SFT), and finally perform reinforcement learning from human feedback (RLHF). 
We evaluate all models on the Chinese medical benchmark datasets, i.e., CMMLU \cite{li-etal-2024-cmmlu} and CEval \cite{huang2023c}.
Specifically, we use the medical-related subsets, namely, Genetics (Gen) and College Medicine (CM) subsets of CMMLU, and the physician (Phy) subset of CEval.
All datasets are multiple-choice questions.
For each question, we prompt the model with the same template and record its top prediction.

Since pre-trained text representation plays an important role in promising text modeling \cite{han2018hyperdoc2vec,ijcai2018-607,devlin-etal-2019-bert,song2021zen,ouyang2022training,touvron2023llama,taori2023alpaca}, we test two pre-trained LLMs in different sizes, namely, Qwen-3 (1.7B)\footnote{\url{https://huggingface.co/Qwen/Qwen3-1.7B}} and Qwen-3 (14B)\footnote{\url{https://huggingface.co/Qwen/Qwen3-14B}} \cite{yang2025qwen3}, following their default settings.
Specifically, Qwen-3 (1.7B) contains 28 layers of self-attention with 2,048-dimensional hidden vectors; and Qwen-3 (14B) contains 40 layers of self-attention with 5,120-dimensional hidden vectors.
During pre-training, each model is trained on the full dataset for two epochs. 
In the SFT stage, we fine-tune on the generated QA pairs for two epochs. 
In the RLHF stage, we further train using the preference data for one epoch.
For other hyperparameters, we report them in Table \ref{tab:hyperparameters}.
We utilize LoRA \cite{hu2022lora} for efficient training.
We evaluate all models based on their accuracy using the LM Evaluation Harness toolkit \cite{gao2024evalharness}\footnote{\url{https://github.com/EleutherAI/lm-evaluation-harness}}.
All models are trained on eight NVIDIA A40 GPUs with 48GB of memory.

\begin{table}[t]
\centering
\small
\scalebox{0.95}{
\begin{tabularx}{0.5\textwidth}{l|l|cc|c}
\toprule
& & \multicolumn{2}{c|}{CMMLU} & CEval  \\
& & Gen & CM & Phy \\
\midrule
\multirow{2}{*}{Qwen-3 (1.7B)} & Original 
& 48.30 & 60.07 & 59.18 
\\
& Ours 
& \textbf{51.14} & \textbf{61.17} & \textbf{63.27}
\\
\midrule
\multirow{2}{*}{Qwen-3 (14B)} & Original 
& 87.55 & \textbf{76.70} & 85.71 
\\
& Ours 
& \textbf{89.01} & \textbf{76.70} & \textbf{87.76}
\\
\bottomrule
\end{tabularx}
}
\vspace{-0.2cm}
\caption{
\textcolor{black}{
The results of different models on medical-related subsets of CMMLU and CEval.
For CMMLU, ``Gen'' and `CM'' stand for genetics and college medicine, respectively.
For CEval, ``phy'' stands for the physician subset.
}
}
\label{tab:overall-results}
\end{table}

\subsection{Results}

\textcolor{black}{
The results of the vanilla LLMs and the ones trained on ChiMed 2.0 with Qwen-3 (1.7B) and Qwen-3 (14B) are presented Table \ref{tab:overall-results}.
Our dataset consistently improves performance across model scales.
On Qwen-3 (1.7B), we observe absolute gains of 2.84\%, 1.10\%, and 4.09\% on the CMMLU Genetics, CMMLU College Medicine, and CEval Physics subsets, respectively.
And for Qwen-3 (14B), our approach yields improvements of 1.46\% and 2.05\% on the CMMLU Genetics and CEval Physics subsets, confirming its effectiveness at larger scales. 
The higher relative gains on smaller models, along with their lower computational requirements, are particularly advantageous for resource-constrained deployments, especially for the medical domain.
Thus, our dataset not only demonstrates broad applicability in academic benchmarks but also offers a practical solution for efficient deployment in healthcare scenarios.
}

\section{Conclusion}

\textcolor{black}{
In this paper, we introduce and construct a large-scale, comprehensive Chinese medical dataset named ChiMed 2.0, which covers pre-training, supervised fine-tuning, and reinforcement learning from human feedback (RLHF), integrating both TCM classics and modern general medical texts.
We design a multi-stage data processing pipeline, including basic cleaning, sensitive content screening, QA pair generation, automated ancient text translation, and preference data generation, to ensure corpus quality and diversity.
In total, the dataset comprises 204.4M characters, supporting various medical NLP downstream tasks such as diagnostic assistance, clinical decision support, and question answering.
Our experiments demonstrate that pre-training, fine-tuning, and RLHF on ChiMed 2.0 yield substantial LLM performance improvement on the medical domain, validating the dataset’s effectiveness and generality.
In future work, we plan to further expand ChiMed-2.0 by covering more clinical specialties and enlarging the preference dataset with additional high‑quality feedback. 
We also aim to incorporate multimodal medical data, such as radiology images and structured electronic health records, to enrich the corpus and support broader AI applications.
}

\bibliography{tacl2021}

\begin{thebibliography}{41}
\expandafter\ifx\csname natexlab\endcsname\relax\def\natexlab#1{#1}\fi

\bibitem[{Abacha et~al.(2019)Abacha, Mrabet, Sharp, Goodwin, Shooshan, and Demner-Fushman}]{abacha2019bridging}
Asma~Ben Abacha, Yassine Mrabet, Mark Sharp, Travis~R Goodwin, Sonya~E Shooshan, and Dina Demner-Fushman. 2019.
\newblock Bridging the gap between consumers’ medication questions and trusted answers.
\newblock In \emph{MEDINFO 2019: Health and Wellbeing e-Networks for All}, pages 25--29. IOS Press.

\bibitem[{Abbasiantaeb et~al.(2024)Abbasiantaeb, Yuan, Kanoulas, and Aliannejadi}]{abbasiantaeb2024let}
Zahra Abbasiantaeb, Yifei Yuan, Evangelos Kanoulas, and Mohammad Aliannejadi. 2024.
\newblock Let the llms talk: Simulating human-to-human conversational qa via zero-shot llm-to-llm interactions.
\newblock In \emph{Proceedings of the 17th ACM International Conference on Web Search and Data Mining}, pages 8--17.

\bibitem[{Chen et~al.(2025)Chen, Fang, Singla, and Dredze}]{chen2025benchmarking}
Hanjie Chen, Zhouxiang Fang, Yash Singla, and Mark Dredze. 2025.
\newblock Benchmarking large language models on answering and explaining challenging medical questions.
\newblock In \emph{Proceedings of the 2025 Conference of the Nations of the Americas Chapter of the Association for Computational Linguistics: Human Language Technologies (Volume 1: Long Papers)}, pages 3563--3599.

\bibitem[{Cui and Han(2020)}]{cui2020chinese}
Xiongtao Cui and Jungang Han. 2020.
\newblock Chinese medical question answer matching based on interactive sentence representation learning.
\newblock \emph{arXiv preprint arXiv:2011.13573}.

\bibitem[{Dai et~al.(2025)Dai, Ye, Liu, Tang, and Zhan}]{dai2025zeus}
Siyuan Dai, Kai Ye, Guodong Liu, Haoteng Tang, and Liang Zhan. 2025.
\newblock Zeus: Zero-shot llm instruction for union segmentation in multimodal medical imaging.
\newblock \emph{arXiv preprint arXiv:2504.07336}.

\bibitem[{Devlin et~al.(2019)Devlin, Chang, Lee, and Toutanova}]{devlin-etal-2019-bert}
Jacob Devlin, Ming-Wei Chang, Kenton Lee, and Kristina Toutanova. 2019.
\newblock {BERT}: Pre-training of deep bidirectional transformers for language understanding.
\newblock In \emph{NAACL}, pages 4171--4186, Minneapolis, Minnesota.

\bibitem[{Gao et~al.(2024)Gao, Tow, Abbasi, Biderman, Black, DiPofi, Foster, Golding, Hsu, Le~Noac'h, Li, McDonell, Muennighoff, Ociepa, Phang, Reynolds, Schoelkopf, Skowron, Sutawika, Tang, Thite, Wang, Wang, and Zou}]{gao2024evalharness}
Leo Gao, Jonathan Tow, Baber Abbasi, Stella Biderman, Sid Black, Anthony DiPofi, Charles Foster, Laurence Golding, Jeffrey Hsu, Alain Le~Noac'h, Haonan Li, Kyle McDonell, Niklas Muennighoff, Chris Ociepa, Jason Phang, Laria Reynolds, Hailey Schoelkopf, Aviya Skowron, Lintang Sutawika, Eric Tang, Anish Thite, Ben Wang, Kevin Wang, and Andy Zou. 2024.
\newblock The language model evaluation harness.

\bibitem[{Guan et~al.(2020)Guan, Zan, Zhou, Xu, and Zhang}]{guan2020cmeie}
Tongfeng Guan, Hongying Zan, Xiabing Zhou, Hongfei Xu, and Kunli Zhang. 2020.
\newblock Cmeie: Construction and evaluation of chinese medical information extraction dataset.
\newblock In \emph{Natural Language Processing and Chinese Computing: 9th CCF International Conference, NLPCC 2020, Zhengzhou, China, October 14--18, 2020, Proceedings, Part I 9}, pages 270--282. Springer.

\bibitem[{Han et~al.(2018)Han, Song, Zhao, Shi, and Zhang}]{han2018hyperdoc2vec}
Jialong Han, Yan Song, Wayne~Xin Zhao, Shuming Shi, and Haisong Zhang. 2018.
\newblock hyperdoc2vec: Distributed representations of hypertext documents.
\newblock \emph{arXiv preprint arXiv:1805.03793}.

\bibitem[{Hu et~al.(2022)Hu, Shen, Wallis, Allen-Zhu, Li, Wang, Wang, Chen et~al.}]{hu2022lora}
Edward~J Hu, Yelong Shen, Phillip Wallis, Zeyuan Allen-Zhu, Yuanzhi Li, Shean Wang, Lu~Wang, Weizhu Chen, et~al. 2022.
\newblock Lora: Low-rank adaptation of large language models.
\newblock \emph{ICLR}, 1(2):3.

\bibitem[{Huang et~al.(2023)Huang, Bai, Zhu, Zhang, Zhang, Su, Liu, Lv, Zhang, Fu et~al.}]{huang2023c}
Yuzhen Huang, Yuzhuo Bai, Zhihao Zhu, Junlei Zhang, Jinghan Zhang, Tangjun Su, Junteng Liu, Chuancheng Lv, Yikai Zhang, Yao Fu, et~al. 2023.
\newblock C-eval: A multi-level multi-discipline chinese evaluation suite for foundation models.
\newblock \emph{Advances in Neural Information Processing Systems}, 36:62991--63010.

\bibitem[{Jin et~al.(2021)Jin, Pan, Oufattole, Weng, Fang, and Szolovits}]{jin2021disease}
Di~Jin, Eileen Pan, Nassim Oufattole, Wei-Hung Weng, Hanyi Fang, and Peter Szolovits. 2021.
\newblock What disease does this patient have? a large-scale open domain question answering dataset from medical exams.
\newblock \emph{Applied Sciences}, 11(14):6421.

\bibitem[{Jin et~al.(2019)Jin, Dhingra, Liu, Cohen, and Lu}]{jin2019pubmedqa}
Qiao Jin, Bhuwan Dhingra, Zhengping Liu, William~W Cohen, and Xinghua Lu. 2019.
\newblock Pubmedqa: A dataset for biomedical research question answering.
\newblock \emph{arXiv preprint arXiv:1909.06146}.

\bibitem[{Johnson et~al.(2016)Johnson, Pollard, Shen, Lehman, Feng, Ghassemi, Moody, Szolovits, Anthony~Celi, and Mark}]{johnson2016mimic}
Alistair~EW Johnson, Tom~J Pollard, Lu~Shen, Li-wei~H Lehman, Mengling Feng, Mohammad Ghassemi, Benjamin Moody, Peter Szolovits, Leo Anthony~Celi, and Roger~G Mark. 2016.
\newblock Mimic-iii, a freely accessible critical care database.
\newblock \emph{Scientific data}, 3(1):1--9.

\bibitem[{Li et~al.(2024{\natexlab{a}})Li, Zhang, Koto, Yang, Zhao, Gong, Duan, and Baldwin}]{li-etal-2024-cmmlu}
Haonan Li, Yixuan Zhang, Fajri Koto, Yifei Yang, Hai Zhao, Yeyun Gong, Nan Duan, and Timothy Baldwin. 2024{\natexlab{a}}.
\newblock {CMMLU}: Measuring massive multitask language understanding in {C}hinese.
\newblock In \emph{Findings of the Association for Computational Linguistics: ACL 2024}, pages 11260--11285, Bangkok, Thailand.

\bibitem[{Li et~al.(2024{\natexlab{b}})Li, Balachandran, Feng, Ilgen, Pierson, Koh, and Tsvetkov}]{li2024mediq}
Shuyue~Stella Li, Vidhisha Balachandran, Shangbin Feng, Jonathan Ilgen, Emma Pierson, Pang~Wei Koh, and Yulia Tsvetkov. 2024{\natexlab{b}}.
\newblock Mediq: Question-asking llms for adaptive and reliable medical reasoning.
\newblock \emph{arXiv e-prints}, pages arXiv--2406.

\bibitem[{Liu et~al.(2023{\natexlab{a}})Liu, Zhou, Hua, Chong, Tian, Liu, Wang, You, Guo, Zhu et~al.}]{liu2023benchmarking}
Junling Liu, Peilin Zhou, Yining Hua, Dading Chong, Zhongyu Tian, Andrew Liu, Helin Wang, Chenyu You, Zhenhua Guo, Lei Zhu, et~al. 2023{\natexlab{a}}.
\newblock Benchmarking large language models on cmexam-a comprehensive chinese medical exam dataset.
\newblock \emph{Advances in Neural Information Processing Systems}, 36:52430--52452.

\bibitem[{Liu et~al.(2023{\natexlab{b}})Liu, Wang, Yang, and Zha}]{liu2023fingpt}
Xiao-Yang Liu, Guoxuan Wang, Hongyang Yang, and Daochen Zha. 2023{\natexlab{b}}.
\newblock Fingpt: Democratizing internet-scale data for financial large language models.
\newblock \emph{arXiv preprint arXiv:2307.10485}.

\bibitem[{Monajatipoor et~al.(2024)Monajatipoor, Yang, Stremmel, Emami, Mohaghegh, Rouhsedaghat, and Chang}]{monajatipoor2024llms}
Masoud Monajatipoor, Jiaxin Yang, Joel Stremmel, Melika Emami, Fazlolah Mohaghegh, Mozhdeh Rouhsedaghat, and Kai-Wei Chang. 2024.
\newblock Llms in biomedicine: A study on clinical named entity recognition.
\newblock \emph{arXiv preprint arXiv:2404.07376}.

\bibitem[{Ouyang et~al.(2022)Ouyang, Wu, Jiang, Almeida, Wainwright, Mishkin, Zhang, Agarwal, Slama, Ray et~al.}]{ouyang2022training}
Long Ouyang, Jeffrey Wu, Xu~Jiang, Diogo Almeida, Carroll Wainwright, Pamela Mishkin, Chong Zhang, Sandhini Agarwal, Katarina Slama, Alex Ray, et~al. 2022.
\newblock Training {L}anguage {M}odels to {F}ollow {I}nstructions with {H}uman {F}eedback.
\newblock \emph{Advances in Neural Information Processing Systems}, 35:27730--27744.

\bibitem[{Pal et~al.(2022)Pal, Umapathi, and Sankarasubbu}]{pal2022medmcqa}
Ankit Pal, Logesh~Kumar Umapathi, and Malaikannan Sankarasubbu. 2022.
\newblock Medmcqa: A large-scale multi-subject multi-choice dataset for medical domain question answering.
\newblock In \emph{Conference on health, inference, and learning}, pages 248--260. PMLR.

\bibitem[{Park et~al.(2023)Park, O'Brien, Cai, Morris, Liang, and Bernstein}]{park2023generative}
Joon~Sung Park, Joseph O'Brien, Carrie~Jun Cai, Meredith~Ringel Morris, Percy Liang, and Michael~S Bernstein. 2023.
\newblock Generative agents: Interactive simulacra of human behavior.
\newblock In \emph{Proceedings of the 36th annual acm symposium on user interface software and technology}, pages 1--22.

\bibitem[{Radford et~al.(2019)Radford, Wu, Child, Luan, Amodei, Sutskever et~al.}]{radford2019language}
Alec Radford, Jeffrey Wu, Rewon Child, David Luan, Dario Amodei, Ilya Sutskever, et~al. 2019.
\newblock Language models are unsupervised multitask learners.
\newblock \emph{OpenAI blog}, 1(8):9.

\bibitem[{Song and Shi(2018)}]{ijcai2018-607}
Yan Song and Shuming Shi. 2018.
\newblock Complementary {L}earning of {W}ord {E}mbeddings.
\newblock In \emph{Proceedings of the Twenty-Seventh International Joint Conference on Artificial Intelligence, {IJCAI-18}}, pages 4368--4374.

\bibitem[{Song et~al.(2020)Song, Tian, Wang, and Xia}]{song-etal-2020-summarizing}
Yan Song, Yuanhe Tian, Nan Wang, and Fei Xia. 2020.
\newblock Summarizing {M}edical {C}onversations via {I}dentifying {I}mportant {U}tterances.
\newblock In \emph{Proceedings of the 28th International Conference on Computational Linguistics}, pages 717--729.

\bibitem[{Song et~al.(2021)Song, Zhang, Wang, and Lee}]{song2021zen}
Yan Song, Tong Zhang, Yonggang Wang, and Kai-Fu Lee. 2021.
\newblock {ZEN} 2.0: {C}ontinue {T}raining and {A}daption for {N}-gram {E}nhanced {T}ext {E}ncoders.
\newblock \emph{arXiv preprint arXiv:2105.01279}.

\bibitem[{Taori et~al.(2023)Taori, Gulrajani, Zhang, Dubois, Li, Guestrin, Liang, and Hashimoto}]{taori2023alpaca}
Rohan Taori, Ishaan Gulrajani, Tianyi Zhang, Yann Dubois, Xuechen Li, Carlos Guestrin, Percy Liang, and Tatsunori~B. Hashimoto. 2023.
\newblock {Stanford Alpaca: An Instruction-following LLaMA model}.
\newblock \emph{GitHub repository}.

\bibitem[{Tian et~al.(2024{\natexlab{a}})Tian, Gan, Song, Zhang, and Zhang}]{tian-etal-2024-chimed}
Yuanhe Tian, Ruyi Gan, Yan Song, Jiaxing Zhang, and Yongdong Zhang. 2024{\natexlab{a}}.
\newblock {C}hi{M}ed-{GPT}: A {C}hinese medical large language model with full training regime and better alignment to human preferences.
\newblock In \emph{Proceedings of the 62nd Annual Meeting of the Association for Computational Linguistics (Volume 1: Long Papers)}, pages 7156--7173, Bangkok, Thailand.

\bibitem[{Tian et~al.(2019)Tian, Ma, Xia, and Song}]{tian2019chimed}
Yuanhe Tian, Weicheng Ma, Fei Xia, and Yan Song. 2019.
\newblock Chimed: A chinese medical corpus for question answering.
\newblock In \emph{Proceedings of the 18th BioNLP Workshop and Shared Task}, pages 250--260.

\bibitem[{Tian et~al.(2025)Tian, Su, Duan, and Song}]{tian2025computed}
Yuanhe Tian, Chen Su, Junwen Duan, and Yan Song. 2025.
\newblock Computed tomography visual question answering with cross-modal feature graphing.
\newblock \emph{arXiv preprint arXiv:2507.04333}.

\bibitem[{Tian et~al.(2024{\natexlab{b}})Tian, Xia, and Song}]{tian2024dialogue}
Yuanhe Tian, Fei Xia, and Yan Song. 2024{\natexlab{b}}.
\newblock Dialogue summarization with mixture of experts based on large language models.
\newblock In \emph{Proceedings of the 62nd Annual Meeting of the Association for Computational Linguistics (Volume 1: Long Papers)}, pages 7143--7155.

\bibitem[{Tian et~al.(2024{\natexlab{c}})Tian, Xia, and Song}]{tian2024diffusion}
Yuanhe Tian, Fei Xia, and Yan Song. 2024{\natexlab{c}}.
\newblock Diffusion networks with task-specific noise control for radiology report generation.
\newblock In \emph{Proceedings of the 32nd ACM International Conference on Multimedia}, pages 1771--1780.

\bibitem[{Touvron et~al.(2023)Touvron, Lavril, Izacard, Martinet, Lachaux, Lacroix, Rozi{\`e}re, Goyal, Hambro, Azhar et~al.}]{touvron2023llama}
Hugo Touvron, Thibaut Lavril, Gautier Izacard, Xavier Martinet, Marie-Anne Lachaux, Timoth{\'e}e Lacroix, Baptiste Rozi{\`e}re, Naman Goyal, Eric Hambro, Faisal Azhar, et~al. 2023.
\newblock {LLaMA: Open and Efficient Foundation Language Models}.
\newblock \emph{arXiv preprint arXiv:2302.13971}.

\bibitem[{Wang et~al.(2024)Wang, Ran, Tang, Chang, Chuang, Liu, Braverman, Liu, and Hu}]{wang2024assessing}
Guanchu Wang, Junhao Ran, Ruixiang Tang, Chia-Yuan Chang, Yu-Neng Chuang, Zirui Liu, Vladimir Braverman, Zhandong Liu, and Xia Hu. 2024.
\newblock Assessing and enhancing large language models in rare disease question-answering.
\newblock \emph{arXiv preprint arXiv:2408.08422}.

\bibitem[{Wang et~al.(2023)Wang, Chen, Song, Zhang, Chen, Xiao, Jiang, Li, Wan, Wang et~al.}]{wang2023cmb}
Xidong Wang, Guiming~Hardy Chen, Dingjie Song, Zhiyi Zhang, Zhihong Chen, Qingying Xiao, Feng Jiang, Jianquan Li, Xiang Wan, Benyou Wang, et~al. 2023.
\newblock Cmb: A comprehensive medical benchmark in chinese.
\newblock \emph{arXiv preprint arXiv:2308.08833}.

\bibitem[{Yang et~al.(2025)Yang, Li, Yang, Zhang, Hui, Zheng, Yu, Gao, Huang, Lv et~al.}]{yang2025qwen3}
An~Yang, Anfeng Li, Baosong Yang, Beichen Zhang, Binyuan Hui, Bo~Zheng, Bowen Yu, Chang Gao, Chengen Huang, Chenxu Lv, et~al. 2025.
\newblock Qwen3 technical report.
\newblock \emph{arXiv preprint arXiv:2505.09388}.

\bibitem[{Yang et~al.(2024)Yang, Zhao, Zhu, Zhou, Xu, Jia, and Zan}]{yang2024zhongjing}
Songhua Yang, Hanjie Zhao, Senbin Zhu, Guangyu Zhou, Hongfei Xu, Yuxiang Jia, and Hongying Zan. 2024.
\newblock Zhongjing: Enhancing the chinese medical capabilities of large language model through expert feedback and real-world multi-turn dialogue.
\newblock In \emph{Proceedings of the AAAI conference on artificial intelligence}, volume~38, pages 19368--19376.

\bibitem[{Yuan et~al.(2024)Yuan, Rastogi, Naik, Rajagopal, Goyal, Zhao, Chintagunta, and Ward}]{yuan2024continued}
Dong Yuan, Eti Rastogi, Gautam Naik, Sree~Prasanna Rajagopal, Sagar Goyal, Fen Zhao, Bharath Chintagunta, and Jeff Ward. 2024.
\newblock A continued pretrained llm approach for automatic medical note generation.
\newblock \emph{arXiv preprint arXiv:2403.09057}.

\bibitem[{Zeng et~al.(2020)Zeng, Yang, Ju, Yang, Wang, Zhang, Zhou, Zeng, Dong, Zhang, Fang, Zhu, Chen, and Xie}]{zeng-etal-2020-meddialog}
Guangtao Zeng, Wenmian Yang, Zeqian Ju, Yue Yang, Sicheng Wang, Ruisi Zhang, Meng Zhou, Jiaqi Zeng, Xiangyu Dong, Ruoyu Zhang, Hongchao Fang, Penghui Zhu, Shu Chen, and Pengtao Xie. 2020.
\newblock {M}ed{D}ialog: Large-scale medical dialogue datasets.
\newblock In \emph{Proceedings of the 2020 Conference on Empirical Methods in Natural Language Processing (EMNLP)}, pages 9241--9250, Online.

\bibitem[{Zhao et~al.(2023)Zhao, Lee, and Hsu}]{zhao2023large}
Zirui Zhao, Wee~Sun Lee, and David Hsu. 2023.
\newblock {Large Language Models as Commonsense Knowledge for Large-Scale Task Planning}.
\newblock \emph{arXiv preprint arXiv:2305.14078}.

\bibitem[{Zhou et~al.(2024)Zhou, Shi, Song, Yang, Jin, Guo, and Li}]{zhou2024lawgpt}
Zhi Zhou, Jiang-Xin Shi, Peng-Xiao Song, Xiao-Wen Yang, Yi-Xuan Jin, Lan-Zhe Guo, and Yu-Feng Li. 2024.
\newblock Lawgpt: A chinese legal knowledge-enhanced large language model.
\newblock \emph{arXiv preprint arXiv:2406.04614}.

\end{thebibliography}
\bibliographystyle{acl_natbib}


\onecolumn

\appendix

\end{document}